\documentclass{article}

 \usepackage[final]{nips_2017}

\usepackage[utf8]{inputenc} % allow utf-8 input
\usepackage[T1]{fontenc}    % use 8-bit T1 fonts
\usepackage{hyperref}       % hyperlinks
\usepackage{url}            % simple URL typesetting
\usepackage{booktabs}       % professional-quality tables
\usepackage{amsfonts}       % blackboard math symbols
\usepackage{nicefrac}       % compact symbols for 1/2, etc.
\usepackage{microtype}      % microtypography
\usepackage{graphicx}
\usepackage{subcaption}
\usepackage{amsthm}
\newtheorem{theorem}{Theorem}
\newtheorem{definition}{Definition}
\usepackage{natbib}
\usepackage{amsmath}
\usepackage{algorithm}% http://ctan.org/pkg/algorithm
\usepackage{algpseudocode}% http://ctan.org/pkg/algorithmicx

\title{Robust Deep Reinforcement Learning with Adversarial Attacks}

\author{
  Anay Pattanaik\\
  Coordinated Science Laboratory\\
 University of Illinois at Urbana-Champaign\\
  \texttt{anayp2@illinois.edu} \\
  \And
   Zhenyi Tang\thanks{Equal contribution} \\
 Computer Engineering\\
  University of Illinois at Urbana-Champaign\\
   \texttt{ztang11@illinois.edu} \\
   \And
   Shuijing Liu* \\
  Computer Engineering\\
  University of Illinois at Urbana-Champaign\\
  \texttt{sliu105@illinois.edu} \\
  \And
   Gautham Bommannan \\
  Computer Engineering\\
  University of Illinois at Urbana-Champaign\\
  \texttt{bommnnn2@illinois.edu} \\
  \And
  Girish Chowdhary\\
  Coordinated Science Laboratory\\
  University of Illinois at Urbana-Champaign\\
  \texttt{girishc@illinois.edu} \\
}

\begin{document}

\maketitle

\begin{abstract}
	This paper proposes adversarial attacks for Reinforcement Learning (RL) and then improves the robustness of Deep Reinforcement Learning algorithms (DRL) to parameter uncertainties with the help of these attacks. We show that even a naively engineered attack successfully degrades the performance of DRL algorithm. We further improve the attack using gradient information of an engineered loss function which leads to further degradation in performance. These attacks are then leveraged during training to improve the robustness of RL within robust control framework. We show that this adversarial training of DRL algorithms like Deep Double Q learning and Deep Deterministic Policy Gradients leads to significant increase in robustness to parameter variations for RL benchmarks such as Cart-pole, Mountain Car, Hopper and Half Cheetah environment.
\end{abstract}

\section{Introduction} \label{sec:intro}

%Reinforcement Learning (RL) provides us an interesting alternative to control systems which is primarily based on domain specific models of the plants. RL enables us to deal with systems for which modeling is hard and finding control policies is difficult. However, this comes at the cost of curse of dimensionality. Most of the real world robots have continuous state and are high dimensional.  Thus, RL requires feature engineering to be of any use and most of these features are hand crafted. 
Advances in Deep Neural Networks (DNN) has a tremendous impact in addressing the curse of dimensionality in RL and offers state of the art results in several RL tasks  (\cite{levine2016end}, \cite{schulman2015trust}, \cite{lillicrap2015continuous}, \cite{mnih2015human}, \cite{silver2016mastering}).
%Deep Reinforcement Learning (DRL) offers state of the art results in several RL tasks  (\cite{levine2016end}, \cite{schulman2015trust}, \cite{lillicrap2015continuous}, \cite{mnih2015human}, \cite{silver2016mastering}).
However, it has been shown in \cite{goodfellow2014explaining} that DNN can be fooled easily into predicting wrong label by perturbing the input with adversarial attacks. It opens up interesting frontier regarding robustness of machine learning algorithms in general. Robustness assumes greater importance in the context of robotics and safety critical systems where such adversarial noise may lead to undesirable and hazardous situations. Learning robust and high performance policies for continuous state-action reinforcement learning domains is critical to enable succesful adoption of Deep Reinforcement Learning (DRL) in robotics, autonomy, and control problems. More specifically, robustness to real world parameter variations, such as changes in the weight, friction, or other environmental parameters of the dynamical system are critical.

We address these challenges in an adversarial training framework. We first engineer ``optimal" attack on the DRL agent and then leverage these attacks during training that leads to significant improvement in robustness and improves policy performance in challenging continuous domains. Our approach is loosely inspired from the idea of robust control, in which the best case policy is sought over the set containing the worst possible parameters of the system. We translate this into a problem of best performing policy trained in presence of adversary. The key difference, however, is that while robust control approaches tend to be conservative, our approach leverages the inherent optimization mechanisms in DRL to enable learning of policies that have even higher performance over a range of parameter and dynamical uncertainties like friction, mass etc.

%An interesting avenue is to see whether we can engineer such attacks on reinforcement learning algorithms.
%%  Prior work \cite{huang2017adversarial} on adversarial attack on DRL is similar in spirit to \cite{goodfellow2014explaining} but it is still not clear what optimal attacks are in RL setting.
%Our paper tries to seek answers to this question and propose optimal attacks specific to Reinforcement Learning algorithms. We also attempt to attack Reinforcement learning algorithms that use radial basis function (RBF) as the underlying function approximator. Interestingly, we show that RBFs are relatively robust to adversarial attacks in comparison to DRL. As we shall see in related work section \ref{subsec:related_work}, all prior work on adversarial attack on DRL have focused on environments involving images \cite{} and are similar in spirit to adversarial attack on images. We engineer attacks that are more effective for reinforcement learning algorithms and show that they are better than previously
%engineered attacks.
%
%Another interesting frontier to explore is robustness of DRL algorithms to model variations. RL algorithms can be trained in simulator before being deployed in real world. Absence of high fidelity simulator can adversely affect the performance of these algorithms in real world. Thus, this provided another basis for our work,that is, how can we make DRL algorithms more robust to parameter mismatch between real dynamics and simulated dynamics.

In particular, the contributions of this paper are two-fold: First, we propose an objective function specifically in a reinforcement learning setting whose optimization degrades the performance of RL algorithms. Furthermore, we show that an attack designed to utilize the minima of our objective function ensures that the RL agent subject to the attack is fooled into thinking that it is in a state that leads to the worst possible action for its current state. This in itself is interesting and exposes critical robustness issues with prevalent DRL algorithms. These attacks severely degrade the performance during evaluation of RL policies. Interestingly, we observe that linearly parameterized RL algorithms are more robust to such adversarial attacks as compared to DRL.  Our second contribution is improving the robustness of DRL algorithm to parameter uncertainties. We train the DRL with proposed engineered adversarial attack and show that it becomes robust to parameter or model uncertainties, that is, it shows significant improvement in performance across a wide range of parameters. Specifically, we train the DRL adversarially on "default" parameters, then this adversarially trained agent is tested on a wide range of parameters (and performs much better than baseline). 

The paper is organized as follows. We provide introduction and related work that has been done in Section \ref{sec:intro}. The background has been discussed in Section \ref{sec:background}. Adversarial attacks and their use for improving robustness have been described in Section \ref{sec:method} and results have been presented in Section \ref{sec:results}. Finally, concluding remarks and future directions have been discussed in Section \ref{sec:conclusion}.

%We also examine robustness to such attacks for different function approximators namely deep neural network and radial basis function network. We have shown that for different RL algorithms, RBF is significantly more robust than DNN for these naively generated attacks. 

\subsection{Related Work}\label{subsec:related_work}

In \cite{huang2017adversarial}, interesting results have been presented regarding adversarial attack on reinforcement learning algorithms. The adversarial attacks in \cite{huang2017adversarial} are similar in spirit to fast signed gradient Method (FSGM) (\cite{goodfellow2014explaining}). In \cite{huang2017adversarial}, the probability of an image being classified with a label is replaced by the probability of taking an action as in \cite{goodfellow2014explaining} and results were generated using RL algorithms for Atari game environment (using image observation).  As we shall show in Section \ref{subsec:objective_adv}, the loss function presented in this paper is guaranteed to maximize the probability of taking the worst possible action which is not necessarily the case in \cite{huang2017adversarial}. In \cite{huang2017adversarial}, the adversarial noise is generated without being certain (when the noise is generated) that it indeed causes adversarial attack. In contrast, we exploit the value function to ascertain the efficacy of our attack. In \cite{kos2017delving}, the authors have also used FSGM style of attack to cause performance failure in Atari game using Asynchronous Advantage Actor Critic (A3C) (\cite{mnih2016asynchronous}) algorithm. They reduced the frequency of attack by injecting noise only when trained value function is above a certain threshold. The intuition being that the agent should be disrupted only when it is more likely to win the game.  They retrained using adversarial attacks and found that it becomes more resilient to adversarial attacks. \cite{lin2017tactics} proposed attack that is similar in spirit to \cite{carlini2016towards} attack on images. In enchanting based attack, they tried to lure agent in bad states by first using a predictive model of video to try to find a sequence of actions that leads to good state. Then  a random sequence of actions were generated which might to bad state, perturbation are generated so that the actions that agent takes is similar to this random sequence of action.  However, it is interesting to note that the value function (of trained agents) itself contains information about this possible worst sequence of action. Hence, we have used value function instead to find adversarial action rather than trying to follow another random sequence of action.  All these attacks have been performed on trained agents.

An important point regarding prior work on adversarial attacks for DRL (\cite{huang2017adversarial}, \cite{kos2017delving}, \cite{lin2017tactics}) is that all of them have been performed in environments that use images, that is, high dimensional pixel input. It has been argued (\cite{goodfellow2014explaining}) that images are susceptible to attack given the high dimensional space of pixel inputs. Thus, most attacks on RL algorithms are similar in spirit to attacks used in case of images. Our work for RL algorithm is not restricted to high dimensional image. We also point out concurrent work by \cite{mandlekar2017iros}. However, they use a heuristic in objective function (minimize $||u||^2_{2}$ where $u$ is the action or control input) for adversarial attack which does not ensure that the attack is optimal as compared to our proposed objective for adversary that is optimal attack.

In \cite{rajeswaran2016epopt}, an ensemble of models was for robust reinforcement learning. They sample model parameters and perform trajectory rollout with those parameter variations. Further, parameters are selected to be trained with based on worst performing percentile criteria. However, sampling trajectories with uncertain parameters can be risky. Moreover, sampling may be very difficult for a large number of parameters. Another approach to robust reinforcement learning  has been provided in  \cite{morimoto2005robust} and it has been extended with deep network in in \cite{pinto2017robust}. Their objective was also to sample worst performing percentile trajectories. To achieve this,  adversary and RL agent are trained alternatively with an expectation that the agent becomes robust to adversary. This is similar to max-min formulation of robust control. However, finding equilibrium of max-min formulation can be difficult. Our work provides a direct approach for sampling worst performing trajectories wherein the adversary fools the agent into believing that it is in states which lead the agent into taking bad actions leading to sampling of worst performing trajectories. In concurrent work, \cite{mandlekar2017iros} have proposed adversarial perturbation of states and robust training using this perturbation. However, the objective function that adversary needs to optimize is heuristic and is given by ($||u||^2_{2}$) where $u$ is the control input. They have reported high variance and the results that also have been reported in \cite{mandlekar2017iros} is comparison of highest return over several random agents as opposed to averaging out the returns for all the agents. 

Another relevant body of work in reinforcement learning is the risk sensitive reinforcement learning where the reward is augmented with risk. Risk can be seen as variance of long-term return. Here, higher variance implies more instability and hence, greater risk. The various risk criterion used in literature are variance penalized (\cite{gosavi2014variance}), weighted risk and return criterion (\cite{Geibel:2005:RRL:1622519.1622522}) etc. A survey of risk sensitive RL can be found in \cite{garcia2015comprehensive}. However, these methods do not scale well with respect to space and action complexity as explained in \cite{garcia2015comprehensive}.

\section{Background} \label{sec:background}
We briefly review the adversarial attacks on images (\cite{goodfellow2014explaining}) because they have been extended to image based DRL (\cite{huang2017adversarial}, \cite{kos2017delving}, \cite{lin2017tactics}). We also review some of the algorithms of DRL that have been used in this paper such as Deep Q Learning (DQN), Double Deep Q Learning (DDQN), and Deep Deterministic Policy Gradient (DDPG).
\subsection{Adversarial attack on Deep Learning Network classifiers (Fast Signed Gradient Method)}

One of the most popular ways to engineer adversarial attacks on deep learning classifiers (that have been extended to DRL) is fast signed gradient method (FSGM) (\cite{goodfellow2014explaining}) where a cost function is crafted whose optimization leads to increase in the probability of the network classifying a given image with a wrong label. It takes into account a linear approximation of deep learning model and engineers an attack. Assuming a linear model, $f(x)=w^{T}x$ ($x$ represents the input and $f(x)$ represents the output), the change in output due to perturbation of input by an amount $\eta$ is given by $\tilde{f}(x) = w^{T}x + w^{T}\eta$. We can get maximal perturbation in prediction  with $$\eta_{min} = \epsilon sign (w)$$ Here $\eta_{min}$ represents the best possible adversarial perturbation which is $l_{\infty}$ norm constrained ($\epsilon$). 
However, this attack has also been extended to nonlinear in parameter functions (multi-layer deep neural networks) and has successfully fooled the classifier networks into confident misclassification of images. The adversarial attack is $$\eta_{min} = \epsilon sign (\nabla_{x} J(\theta, x,y))$$ Here, $J(\theta, x,y)$ is the respective loss function that is used during training or testing. Again with the underlying assumption that adversarial noise is $l_{\infty}$ norm bounded. In case of images, $J(\theta, x,y)$ is the cross entropy loss between the true image label and predicted distribution over labels of image

\subsection{Deep Q learning (DQN) and Deep Double Q Learning (DDQN)}

Deep Q learning (DQN) has achieved superhuman results on Atari games \cite{mnih2015human}. Q learning is a value function based algorithm (we refer to both state-action value function and state-value function as value function). Q values of a state-action pair represent how good the action being contemplated by the agent is for its current observation. The learning agent updates the Q value using temporal difference error and simultaneously acts to maximize its return in the long run. In DQN algorithm, the agent uses a Deep neural network to approximate this Q function. The DQN algorithm achieved stability in training primarily by using experience replay and the use of target network. For experience replay, it stores past state, action, reward, next state sequence and these are used to update the Q network just as in supervised learning with these sequences being picked randomly from memory. This breaks the strong correlation between samples. For this ``supervised'' learning type of update, DQN has another neural network, called target network which provides the Q values for the next state(s). The target network is updated by hard transfer of online weights after a fixed number of iterations. This two network structure led to ``stability'' in training. However, \cite{van2016deep} showed that DQN overestimates the Q values and proposed Double Deep Q Learning (DDQN) where the action selection was still performed by online network but it's value estimate for update was done using the target network. This mitigated the overestimation of value function. 

\subsection{Radial Basis Function based Q learning}

In radial basis function based Q learning (\cite{Geramifard:2013:TLF:2688182.2688183}), the DNN is replaced by RBF with gaussian kernel. Here experience replay and target networks are typically not used. The networks learns through stochastic gradient descent using temporal difference error.

\subsection{Deep Deterministic Policy Gradient (DDPG)}

Deep deterministic policy gradient (DDPG) (\cite{lillicrap2015continuous}) uses both an actor and a critic for learning. The critic evaluates a given policy generated by the actor. The weights of critic and actor are updated by gradient descent. The critic network uses the prior concepts of experience replay and target network for its update. The only difference between these update and that of Deep Q learning is that instead of ``hard'' transfer of weights to target after a fixed number of iterations, there is ``soft'' transfer of weights where the weights of target network are incremented by a very small amount towards the online network. 
%The actor network learns through gradient of the policy.

%\subsection{Radial Basis Function based Actor-Critic}
%
%This is similar to DDPG except that the actor and critic are represented by RBF \cite{grondman2012survey}. Here, experience replay and target network are typically not used. The learning takes place through stochastic gradient descent for both actor and critic.

\section{Method}\label{sec:method}
In this section, we explain our methods for adversarial attack and further use these attacks for significantly improving the robustness and performance of reinforcement learning algorithm.
\subsection{Adversarial Attack}
In this subsection, we concentrate on generating adversarial attacks which will cause a trained RL agent to fail. We assume that the mechanism of the attack is by corrupting the observations of the agent of its current state, fooling it into believing it is in a state that causes it to follow a sub-optimal policy for its actual state. We begin by defining what we mean by an adversarial attack in the context of value function based reinforcement learning algorithms:
\begin{definition}{\label{adv_def}}
	An adversarial attack is any possible perturbation that leads the agent into increased probability of taking ``worst'' possible action in that state. Here, the ``worst'' possible action for a trained RL agent is the action which corresponds to least Q value in that state. 
\end{definition}

It is important to note that Def. \ref{adv_def} is valid only for value function based algorithms, that is, the algorithms that use value function to predict the optimality of actions in a given state. Most of the popular state-of-art algorithms such as A3C and DDPG are value function based.  We shall see the objective function that needs to be optimized for achieving adversarial attacks in the context of reinforcement learning in section \ref{subsec:objective_adv}. There is a subtle  but important difference as compared to image classification wherein attack is considered successful if a given image is classified as any other image. There is no concept of ``worst'' possible image whereas RL agents can have ``worst'' possible action. Another assumption is that the agent has been trained sufficiently well so that Q values are close to optimal. This is true in case of tabular Q learning as training time tends to infinity (but this has not been proved in general for nonlinear function approximators like DNN) and serves as motivation for this assumption.

An important point to note is that throughout the paper we have bounded the adversarial attack noise by $l_2$ norm constraint. It is possible to use other norm constraints as well and \cite{papernot2016distillation} showed that distillation is secure under $l_\infty$. Another point is that the RL environments are scaled so that the state space is normalized between $[0,1]$. The magnitude of attack is also normalized.

\subsubsection{Naive adversarial attack}
First, we propose a naive method of generating adversarial attack. The idea behind this attack is to generate random noise (several times in current state that the agent is in) and add it to current state with hope that one of these noise samples will cause the agent to take ``bad" action, that is, sub-optimal action. The quality of attack can be ascertained by the value function. Algo. \ref{alg:naive} outlines the naive adversarial attack for DDQN.\\   
%We show that a naively engineered attack is good enough to generate adversarial attacks on Deep Reinforcement learning policy.
An important point to note is that the attacks are generated during evaluation phase. The adversarial attack is essentially a search across nearby observation which will cause the agent to take wrong action. For generating adversarial attack on the DRL policies, we sample a noise with finite (small) support. Noise is not generated once during every iteration, rather it is sampled for a number of times every iteration with a search for best adversarial noise ($beta$ distributed noise with shifted 0 mean was used for generic setting but in our experiments the parameters of beta distribution were $(1,1)$ which corresponds to uniform noise with 0 mean). The particular noise that causes least estimate of the value function is selected as adversarial noise. Then this noise is added to the current observation. Algo. \ref{alg:naive} is outlined for naive attack on DDQN.

For naive attack on DDPG, the critic network can be used to ascertain value functions when required and actor network determines the behavior policy to pick action. Thus, the objective function used by adversary in this case is the $Q^*_{critic}(s,a)$, that is, the value function determined by the trained critic network. Algorithm for naive attack on DDPG has been provided in Algo. \ref{alg:naive_ddpg} and is similar to Algo. \ref{alg:naive}.\\

%\begin{algorithm}[t]
%\caption{Naive adversarial attack}
%\label{alg:dpmirl}
%	\begin{algorithmic}[1]		
%		\Procedure{Naive}{$a,b$} 
%		\State Determine optimal action  $a^* =  arg \max_{a} Q(s,a), Q^* = \max_{a} Q(s,a)$
%		\State Sample noise $(n)\sim beta(\alpha, \beta)$
%		\State new state $s' = s + n$
%		\State Determine optimal action in new state  $a_{adv} =  arg \max_{a} Q(s',a)$ 
%		\State Determine the value function of that action in current state: $Q^*_{adv} = Q(s, a^*_{adv})$
%		\If{$Q^*_{adv} < Q^*$}
%		\State $a^*_{adv} =  a_{adv}$
%		\Else
%		\State do nothing
%		\EndIf
%		\State \bfseries{return} $a^*_{adv}$
%	\end{algorithmic}
%\end{algorithm}	

\begin{algorithm}
	\caption{Naive attack (DDQN)}\label{alg:naive}
	\begin{algorithmic}[1]
		\Procedure{Naive}{$Q^{target},Q,s,\epsilon,n,\alpha,\beta$}\Comment{Naive attack function takes Q network $(Q)$, current state(s), adversarial attack magnitude constraint$(\epsilon)$, parameters of beta distribution$(\alpha,\beta)$ and number of times to sample noise$(n)$ as input}
		\State   $a^* =  arg \max\limits_{a} Q(s,a), Q^* = \max\limits_{a} Q^{target}(s,a)$\Comment{Determine optimal action value function}
		\For{$i = 1:n$}\Comment{Sample a few times}
		\State $n_i\sim beta(\alpha, \beta) - 0.5$\Comment{Sample noise}
		\State $s_i = s + \epsilon *n_i$\Comment{Possible adversarial state determined by sampled noise}
		\State $a_{adv} =  arg \max\limits_{a} Q(s_i,a)$ \Comment{Determine optimal action in potential adversarial state}
		\State $Q^{target}_{adv} = Q^{target}(s, a_{adv})$ \Comment{Determine the value of  potential adversarial action corresponding to potential adversarial state for current state} 
		\If{$Q^{target}_{adv} < Q^*$}\Comment{if the potential adversarial state  leads to bad action}
		%		\State $a^*_{adv} =  a_{adv}$\Comment{Store that action as bad action}
		\State $Q^* = Q^{target}_{adv}$\Comment{Store the value function of that potential bad action}
		\State $s_{adv} =  s_i$\Comment{Store possible adversarial state }
		\Else
		\State do nothing
		\EndIf
		\EndFor\label{euclidendwhile}
		\State \textbf{return} $s_{adv}$\Comment{Adversarial state}
		\EndProcedure
	\end{algorithmic}
\end{algorithm}

\begin{algorithm}
	\caption{Naive attack (DDPG)}\label{alg:naive_ddpg}
	\begin{algorithmic}[1]
		\Procedure{Naive}{$Q^{target}, U ,s,\epsilon,n,\alpha,\beta$}\Comment{Naive attack function takes trained target critic network $Q^{target}$, trained actor network $U$, current state(s), adversarial attack magnitude constraint$(\epsilon)$, parameters of beta distribution$(\alpha,\beta)$ and number of times to sample noise$(n)$ as input}
		\State  $a^*=U^(s), Q^* = Q^{target}(s,a^*)$\Comment{Determine optimal action and action value function}
		\For{$i = 1:n$}\Comment{Sample a few times}
		\State $n_i\sim beta(\alpha, \beta) - 0.5$\Comment{Sample noise}
		\State $s_i = s + \epsilon *n_i$\Comment{Possible adversarial state determined by sampled noise}
		\State $a_{adv} =  U(s_i)$ \Comment{Determine optimal action in potential adversarial state}
		\State $Q^{target}_{adv} = Q^{target}(s, a_{adv})$ \Comment{Determine the value of  potential adversarial action corresponding to potential adversarial state for current state} 
		\If{$Q^{target}_{adv} < Q^*$}\Comment{if the potential adversarial state  leads to bad action}
		%		\State $a^*_{adv} =  a_{adv}$\Comment{Store that action as bad action}
		\State $Q^* = Q^{target}_{adv}$\Comment{Store the value function of that potential bad action}
		\State $s_{adv} =  s_i$\Comment{Store possible adversarial state }
		\Else
		\State do nothing
		\EndIf
		\EndFor\label{euclidendwhile}
		\State \textbf{return} $s_{adv}$\Comment{Adversarial state}
		\EndProcedure
	\end{algorithmic}
\end{algorithm}

\subsubsection{Gradient based adversarial attack }\label{subsec:objective_adv}
In this subsection, we show that a proposed cost function different from the one used in traditional FSGM (\cite{huang2017adversarial})) is more effective in finding worst possible action in the context of reinforcement learning with discrete actions.

\begin{theorem}\label{thm:adv_obj}
	Let the optimal policy be given by conditional probability mass function (pmf) $\pi^{*}(a|s)$, the action which has maximum pmf be given as $a^*$ and the worst possible action be given by $a_{w}$. Then the objective function whose minimization leads to optimal adversarial attack on RL agent is given by
	$$ J(s,\pi^{*}) = -\sum_{i=1}^n p_i log \pi^{*}_i $$ where $\pi^{*}_i = \pi^{*}(a_i|s)$, $p_i = P(a_i)$, the adversarial probability distribution P is given by 
	\begin{equation}\label{eq:adv_prob}
	P(a_i)
	=\left\{
	\begin{array}{@{}ll@{}}
	1, & \text{if}\ a_{w}=1 \\
	0, & \text{otherwise}
	\end{array}\right.
	\end{equation}
	In other words, this is the cross entropy loss between the adversarial probability distribution and optimal policy generated by the RL agent
\end{theorem} 
Q values can be converted into pmf by passing them through softmax function.
\proof
We shall show that  $\min\limits_{s} J(s,\pi^{*})$ achieves the objective of def. \ref{adv_def}. 
\begin{align*}
J(s,\pi^{*}) &= -\sum_{i=1}^n p_i log (\pi^{*}_i) \\
&= - p_{w} log(\pi^{*}_w) = -log (\pi^{*}_w) \quad(\text{from} \ref{eq:adv_prob} )\\
\implies\min_{s} J(s,\pi^{*}) &= \min_{s} -log(\pi^{*}_w)\\
\min_{s} J(s,\pi^{*}) &= \max_{s} log(\pi^{*}_w)
\end{align*}
Since log is monotonically increasing,
\begin{align*}
\min_{s} J(s,\pi^{*}) = \max_{s} \pi^{*}_w
\end{align*}

Thus, we have shown that the objective function that should be used for engineering attack on RL algorithm should be given by Theorem \ref{thm:adv_obj} as it is consistent with Def. \ref{adv_def}.
%(use of other divergence metrics apart from KL divergence is interesting and will be examined in future).
FSGM algorithm can be used to minimize this objective function. We must point out that this objective function is different from ones in literature \cite{huang2017adversarial}. The objective functions mentioned in \cite{huang2017adversarial} will result in $\min\limits_{s}\pi^{*}(a^*|s)$ ($a^*$ is the best possible action for given state $s$). This leads to decrease in the probability of taking best possible action. This won't necessarily lead to increase in probability of taking worst possible action. The gradient based attack for DDQN has been explained in Algo. \ref{alg:grad}

The gradient based attack for DDPG is similar to DDQN with the objective  function that adversary need to minimize being given by the optimal value function of critic ($Q^*(s,a)$). Here the gradient is given by
%\begin{align*}
%\nabla_{s}Q^{target}(s,a) = \frac{\partial Q^{target}}{\partial s} + \frac{\partial Q^{target}}{\partial U^{target}}  \frac{\partial U^{target}}{\partial a}
%\end{align*}\\
\begin{align*}
\nabla_{s}Q^*(s,a) = \frac{\partial Q^*}{\partial s} + \frac{\partial Q^*}{\partial U^*}  \frac{\partial U^*}{\partial s}
\end{align*}\\
where $U^*$ represents the optimal policy given by actor. The algorithm has been provided in Algo. \ref{alg:grad_ddpg} and is similar to Algo. \ref{alg:grad}.
%For performing the attack on DDPG, we followed approach similar to naive sampling based attack but this time the search is more directed in the gradient direction of objective function described in Theorem \ref{thm:adv_obj}. The algorithm has been described in Algo. \ref{alg:grad}

%For a reinforcement learning algorithm in continuous action space with value function approximator, the policy is decided by another neural network in actor-critic framework such as DDPG and A3C. Thus, the observation that leads to minimization of Value function can be considered to be good adverserial attack. The objective function here is straight forward, that is, the value function of the state (V(s)).

\begin{algorithm}
	\caption{Gradient based attack (DDQN)}\label{alg:grad}
	\begin{algorithmic}[1]
		\Procedure{Grad}{$Q^{target},Q,s,\epsilon,n,\alpha,\beta$}\Comment{Gradient based attack function takes Q network $(Q)$, current state$(s)$, adversarial attack magnitude constraint$(\epsilon)$, parameters of beta distribution$(\alpha,\beta)$ and number of times to sample noise$(n)$ as input}
		\State   $a^* =  arg \max\limits_{a} Q(s,a), Q^* = \max\limits_{a} Q^{target}(s,a)$\Comment{Determine optimal action and value function}
		\State $\pi^{target} = softmax(Q^{target})$\Comment{Pass Q through softmax layer to convert it into pmf}
		\State $grad = \nabla_{s} J(s,\pi^{target})$\Comment{Determine the gradient} 
		\State $grad\_dir = \frac{\nabla_{s} J(s,\pi^{target})}{||\nabla_{s} J(s,\pi^{target})||}$\Comment{$l_2$ constrained norm of gradient}
		\For{$i = 1:n$}\Comment{Sample a few times}
		\State $n_i\sim beta(\alpha, \beta)$\Comment{Sample noise}
		\State $s_i = s - n_i*grad\_dir$\Comment{Possible adversarial state determined by sampled noise in the direction of gradient}
		\State $a_{adv} =  arg \max\limits_{a} Q(s_i,a)$ \Comment{Determine optimal action in potential adversarial state}
		\State $Q^{target}_{adv} = Q^{target}(s, a_{adv})$ \Comment{Determine the value of  potential adversarial action corresponding to potential adversarial state for current state} 
		\If{$Q^{target}_{adv} < Q^*$}\Comment{if the potential adversarial state  leads to bad action}
		%		\State $a^*_{adv} =  a_{adv}$\Comment{Store that action as bad action}
		\State $Q^* = Q^{target}_{adv}$\Comment{Store the value function of that potential bad action}
		\State $s_{adv} =  s_i$\Comment{Store that state as possible adversarial state}
		\Else
		\State do nothing
		\EndIf
		\EndFor\label{euclidendwhile}
		\State \textbf{return} $s_{adv}$\Comment{Return adversarial state}
		\EndProcedure
	\end{algorithmic}
\end{algorithm}

\begin{algorithm}
	\caption{Gradient based attack (DDPG)}\label{alg:grad_ddpg}
	\begin{algorithmic}[1]
		\Procedure{Grad}{$Q^{target},U,s,\epsilon,n,\alpha,\beta$}\Comment{Gradient based attack function takes target Q network (critic) $Q^{target}$ , actor network $U$,  current state$(s)$, adversarial attack magnitude constraint$(\epsilon)$, parameters of beta distribution$(\alpha,\beta)$ and number of times to sample noise$(n)$ as input}
		\State   $a^* =  U(s), Q^* = Q^{target}(s,a^*)$\Comment{Determine optimal action and value function}
		\State $grad = \nabla_{s} Q^{target}(s,a)$\Comment{Determine the gradient} 
		\State $grad\_dir = \frac{\nabla_{s} Q^{target}(s,a)}{||\nabla_{s} Q^{target}(s,a)||}$\Comment{$l_2$ constrained norm of gradient}
		\For{$i = 1:n$}\Comment{Sample a few times}
		\State $n_i\sim beta(\alpha, \beta)$\Comment{Sample noise}
		\State $s_i = s - n_i*grad\_dir$\Comment{Possible adversarial state determined by sampled noise in the direction of gradient}
		\State $a_{adv} =  U(s_i)$ \Comment{Determine optimal action in potential adversarial state}
		\State $Q^{target}_{adv} = Q^{target}(s, a_{adv})$ \Comment{Determine the value of  potential adversarial action corresponding to potential adversarial state for current state} 
		\If{$Q^{target}_{adv} < Q^*$}\Comment{if the potential adversarial state  leads to bad action}
		%		\State $a^*_{adv} =  a_{adv}$\Comment{Store that action as bad action}
		\State $Q^* = Q^{target}_{adv}$\Comment{Store the value function of that potential bad action}
		\State $s_{adv} =  s_i$\Comment{Store that state as possible adversarial state}
		\Else
		\State do nothing
		\EndIf
		\EndFor\label{euclidendwhile}
		\State \textbf{return} $s_{adv}$\Comment{Return adversarial state}
		\EndProcedure
	\end{algorithmic}
\end{algorithm}

\subsubsection{SGD based attack}
We also used Stochastic Gradient Descent approach for adversarial attack wherein instead of sampling a few times and selecting best possible attack amongst these samples, we followed the gradient descent for same number of sampling time and selected the state that we end up in as adversarial state.

\subsection{Adversarial Training through Robust Control framework}
In this subsection, we propose robust adversarial training and show equivalence to robust control.
\subsubsection{Robust Control Framework}
In RL, the typical objective that an agent seeks to maximize is it's expected long term return $(\eta)$ (over possible trajectories $\tau$) assuming a fixed transition model $T(s_{t}, a_t;\phi)$ characterized by parameters $\phi$
\begin{align*}
\eta(\pi,T) = \mathbb{E_{\tau}}[\sum_{t=0}^T \gamma^{t}r(s_t,a_t)|s_0,\pi,T] 
\end{align*}
However, if there is variation in transition model, then criteria might be to perform well in expectation over all the possible transition models. Thus, leading to optimization of the mean performance of agent. The objective function in this scenario can be modified to
\begin{align*}
\eta(\pi) = \mathbb{E}_T[\eta(\pi,T)] 
\end{align*}
This is popularly known as risk neutral formulation. However, it has an underlying assumption that the distribution over transition model parameters are known apriori. It may not perform well over the transition model distributions because of high variance in transition model distribution. Thus, conditional value of risk (CVaR) can be used as optimization criteria for robust control (\cite{tamar2015optimizing})
\begin{align*}
\eta_{RC}(\pi) = \mathbb{E}_T[\eta(\pi,T)|\mathbb{P}(\eta(\pi,T)\leq\beta)=\alpha] 
\end{align*} 

So, the problem boils down to maximizing the expected return over worst $\alpha$ percentile of returns. Thereafter, for sampling these bad trajectories, \cite{rajeswaran2016epopt} changed transition model parameters and sample trajectories by performing rollouts with different parameters. \cite{morimoto2005robust} and \cite{pinto2017robust} took an indirect approach where instead of sampling worst trajectory from rollout, they employ an adversary which applies control action and tries to push the RL agent into possible bad states. They train the adversary whose reward was negative of the reward of RL agent resulting in max-min game theoretic formulation. But it is usually difficult to find this equilibrium.\\
\subsubsection{Adversarial Training} 
In contrast to approaches given in \cite{rajeswaran2016epopt} and \cite{morimoto2005robust} where it might be difficult to find equilibrium, we take a direct approach where the adversary fools the agent into believing that it's in a ``fooled'' state different from actual state such that the optimal action in ``fooled'' state leads to worst action in actual current state. In other words, the adversary fools the agent into sampling worst trajectories directly. We first train the algorithm using ''vanilla'' DRL (DDQN or DDPG), the trained agent is then made robust to model uncertainties through adversarial training. We have used gradient based attack for adversarial training as it performed best amongst all attacks (results presented in Section \ref{sec:results}). Adversarial training algorithm has been discussed in Algo. \ref{alg:adv_train} (DDQN) and Algo. \ref{alg:adv_train_ddpg} (DDPG). In our approach, worst $\alpha$ percentile of returns is related to the magnitude of adversarial attack, higher adversary magnitude corresponds to optimization for higher $\alpha$ worst percentile.

%So, the optimization problem can be viewed as
%\begin{align*}
%\eta_{RC}^* =\max_{\pi}\min_{s_{adv}}R(s,\pi,\pi_{adv})
%\end{align*}
%where
%\begin{align*}
%R(s,\pi,\pi_{adv}) = \mathbb{E_\tau}[\sum_{t=t'}^T \gamma^{t}r(s_t,a_t)|s_{t'},\pi,\pi_{adv}]
%\end{align*}
%Here $s_{adv}$ is the state induced by adversarial attacker whose policy is denoted by $\pi_{adv}$, $\pi$ is the RL policy that agent is trying to learn. 
%The algorithm has been discussed in Algo. \ref{alg:adv_train}

\begin{algorithm}
	\caption{Training with adversarial perturbation (DDQN)}\label{alg:adv_train}
	\begin{algorithmic}[1]
		\Procedure{Adv train }{$Q^{target}, Q$}\Comment{Gradient based adversarial training method takes pre-trained network} 
		%		\State $V^* = V(s;\theta_v)$ \Comment{Determine optimal value function of current state}
		%%		\State   $a^* =  arg \max\limits_{a} Q(s,a), Q^* = \max\limits_{a} Q(s,a)$\Comment{Determine optimal action and value function}		
		%		\State $grad = \nabla_{s} V(s;\theta_v)$\Comment{Determine the gradient} 
		%		\State $grad\_dir = \frac{\nabla_{s} V(s;\theta_v)}{||\nabla_{s} V(s;\theta_v)||}$\Comment{$l_2$ constrained norm of gradient}
		\For{$i = 1:iterations$}\Comment{Train adversarially for number of timesteps}
		\State Reset the environment and receive observation
		\While{not terminal or not max time steps per episode reached}
		\State $s_{adv} = Grad(Q^{target},Q,s,\epsilon, n, \alpha, \beta)$\Comment{Fool the agent}
		%		\State $a_{adv} =  \pi_{\theta_p}(a|s_i)$ \Comment{Determine optimal action in new state}
		\State $a = arg \max\limits_{a} Q(s_{adv,a})$ \Comment{Fooled agent takes action according to behavior policy} 
		\State $s,r=Env(a,s)$ \Comment{Environment returns next state and reward corresponding to state $s$ and action $a$}
		\State Update the weights of network according to DDQN algorithm
		%		\Comment{Previous steps can be repeated in rollout as required in A3C}
		\EndWhile 
		\EndFor\label{euclidendwhile}
		\EndProcedure
	\end{algorithmic}
\end{algorithm} 
\begin{algorithm}
	\caption{Training with adversarial perturbation (DDPG)}\label{alg:adv_train_ddpg}
	\begin{algorithmic}[1]
		\Procedure{Adv train }{$Q^{target}, Q, U^{target}, U$}\Comment{Gradient based adversarial training method takes pre-trained network} 
		%		\State $V^* = V(s;\theta_v)$ \Comment{Determine optimal value function of current state}
		%%		\State   $a^* =  arg \max\limits_{a} Q(s,a), Q^* = \max\limits_{a} Q(s,a)$\Comment{Determine optimal action and value function}		
		%		\State $grad = \nabla_{s} V(s;\theta_v)$\Comment{Determine the gradient} 
		%		\State $grad\_dir = \frac{\nabla_{s} V(s;\theta_v)}{||\nabla_{s} V(s;\theta_v)||}$\Comment{$l_2$ constrained norm of gradient}
		\For{$i = 1:iterations$}\Comment{Train adversarially for number of timesteps}
		\State Reset the environment and receive observation
		\While{not terminal or not max time steps per episode reached}
		\State $s_{adv} = Grad(Q^{target},U, s,\epsilon, n, \alpha, \beta)$\Comment{Fool the agent}
		%		\State $a_{adv} =  \pi_{\theta_p}(a|s_i)$ \Comment{Determine optimal action in new state}
		\State $a =  U(s_{adv})$ \Comment{Fooled agent takes action according to behavior policy} 
		\State $s,r=Env(a,s)$ \Comment{Environment returns next state and reward corresponding to state $s$ and action $a$}
		\State Update the weights of network according to DDPG algorithm
		%		\Comment{Previous steps can be repeated in rollout as required in A3C}
		\EndWhile 
		\EndFor\label{euclidendwhile}
		\EndProcedure
	\end{algorithmic}
\end{algorithm}
%Robust control refer to a class of control algorithms that are robust to parameter variations, measurement noise. Usually, it is assumed that the bounds on parameter variation is known apriori. However, most of these techniques are limited to linear systems. We take inspiration from game theoretic formulation of robust control. Wherein the objective of controller is to maximize the return for RL agent whereas the adversary tries to minimize the this return. The aproach taken in pinto et al is that this adversary can be viewed as an external control effort which tries to destabilize the RL agent. Here, we take a complimentary approach. We assume that the adversary fools agent into believing that it is in a state such that the agent takes sub-optimal action. When trained with such an adversary, the agent becomes robust to it. Both these approaches are similar in sense that with Markovian assumption of dynamics
%\begin{align*}
%s_{t+1}=T(s_{t}, a_t;\phi)
%\end{align*}. Here $s_{t+1}$ is the next state while $T(s_{t}, a_t)$ represents the transition dynamics and $a_t$ represents the action taken by the RL agent. The parameters on which transition model depend is represented by $\phi$. Hence, uncertainty and variations in $\phi$ will manifest in variation over the next state. This insight forms the basis of our approach. In other words, our approach can be presented as 
%\begin{align*}
%R_{rob} =\max_{\pi}\min_{s_{adv}}R(s,\pi,\pi_{adv})
%\end{align*}
%Here $s_{adv}$ is the state induced by adversarial attacker whose policy is denoted by $\pi_{adv}$

\section{Results}\label{sec:results}
\begin{figure*}
	
	\begin{subfigure}{0.27\textwidth}
		\includegraphics[trim={0 0 0 0},clip,width=\linewidth]{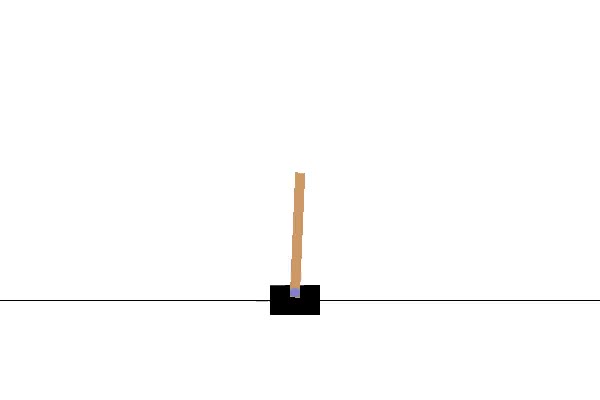} 
		\caption{Cart-Pole}
		%		\label{fig:att_cart}
	\end{subfigure}
	\begin{subfigure}{0.27\textwidth}
		\includegraphics[trim={0 0 0 0},clip,width=\linewidth]{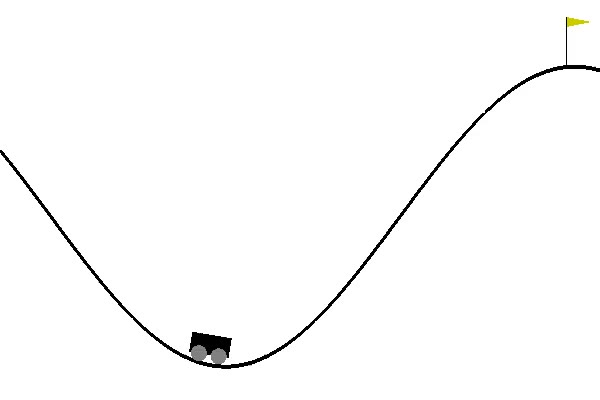}
		\caption{Mountain Car}
		%		\label{fig:rew_mouncar}
	\end{subfigure}	
	\begin{subfigure}{0.2\textwidth}
		\includegraphics[trim={0 0 0 0},clip,width=\linewidth]{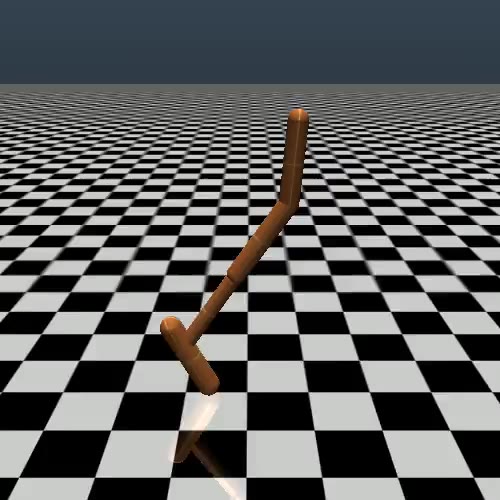}
		\caption{Hopper}
		%		\label{fig:rew_mouncar}
	\end{subfigure}	
	\begin{subfigure}{0.2\textwidth}
		\includegraphics[trim={00 0 0 0},clip,width=\linewidth]{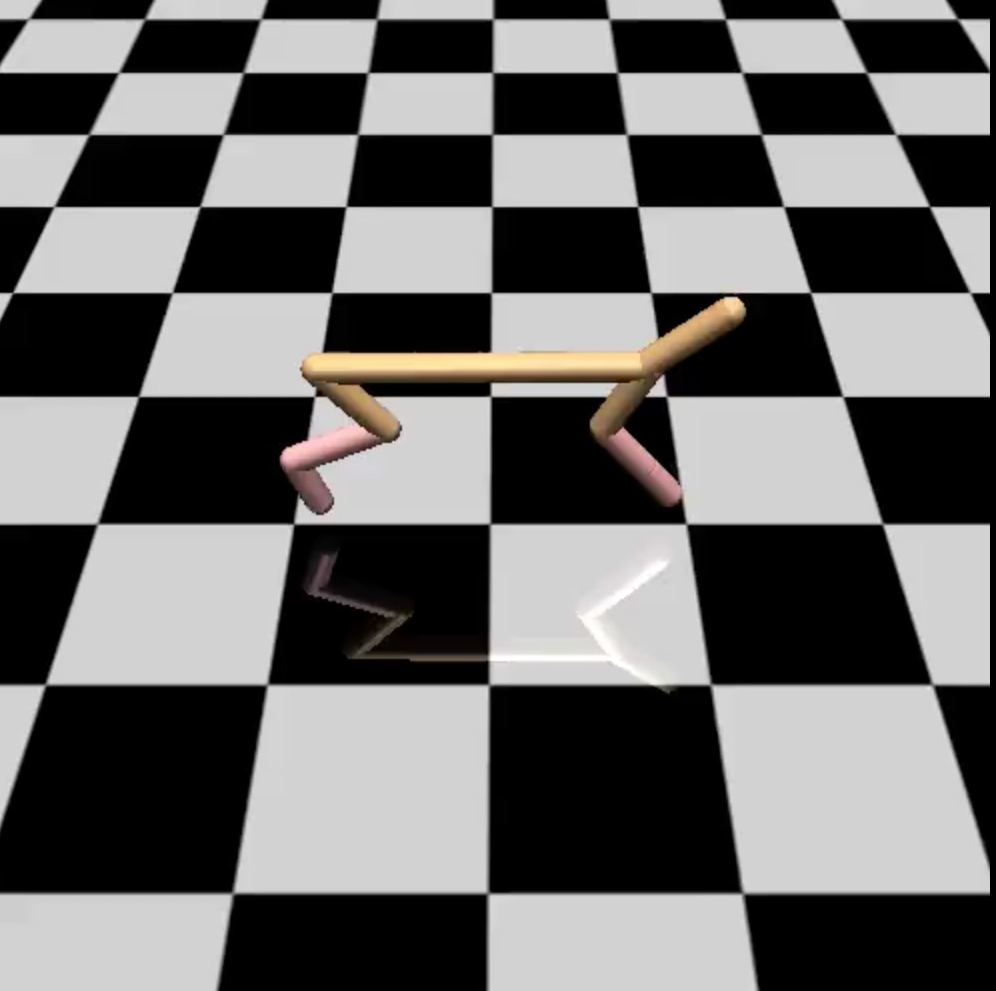}
		\caption{Half Cheetah}
		%		\label{fig:rew_mouncar}
	\end{subfigure}
	\caption{Different environments in OpenAi gym (MuJoCo for simulation of dynamics)
		% It is clear that the naive adversarial attack has significantly degraded the performance of DQN while there is no significant degradation in performance of RBF based Q learning
	}
	\label{fig:gym_env}
\end{figure*}

In this section, we discuss results related to proposed adversarial attack and adversarially trained robust policy. We report improvement to robustness over two algorithms (DDQN and DDPG). All the experiments have been performed within OpenAi gym environment (\cite{brockman2016openai}) with MuJoCo (\cite{todorov2012mujoco}) (Fig. \ref{fig:gym_env})

\vspace{-0.1in}

\begin{figure*}[!h]
	\begin{center}
		\begin{subfigure}{0.24\textwidth}
			\includegraphics[trim={0 5 43 35},clip,width=\linewidth]{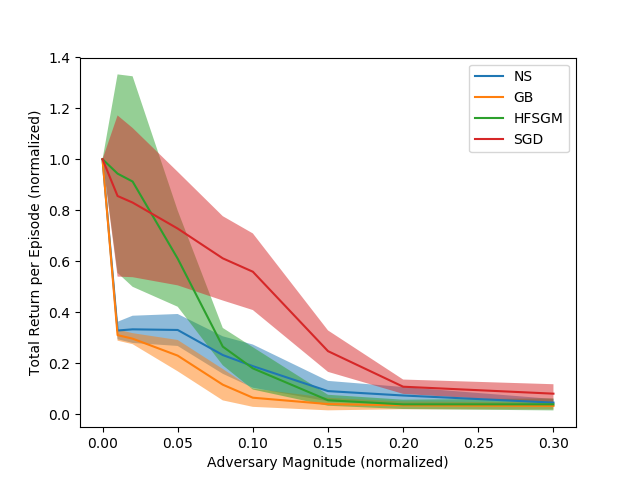} 
			\caption{DDQN Cart Pole}
			%		\label{fig:att_cart}
		\end{subfigure} %\hspace{1 cm}
		\begin{subfigure}{0.24\textwidth}
			\includegraphics[trim={0 5 43 35},clip,width=\linewidth]{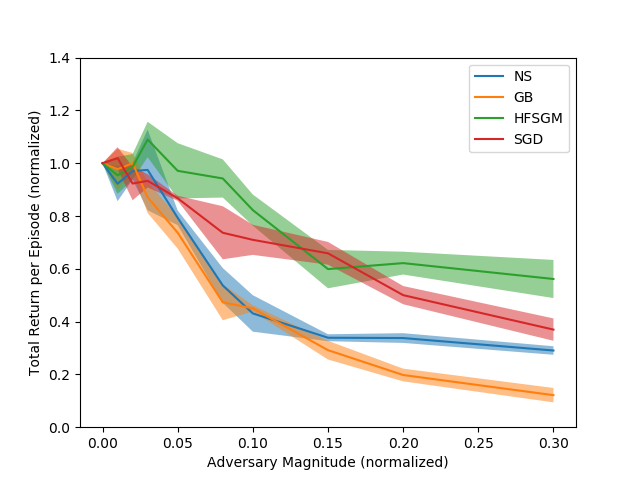}
			\caption{RBF Q Cart Pole}
			%		\label{fig:rew_mouncar}
		\end{subfigure}
		%	\caption{Comparison of different attacks on DDQN and RBF based Q learning for cartpole environment. RBF Q learning is relatively more resilient to adversarial attack.
		% It is clear that the naive adversarial attack has significantly degraded the performance of DQN while there is no significant degradation in performance of RBF based Q learning
		%	}
		\label{fig:att_cartpole}	
		%\end{figure}
		%\begin{figure}[!t]
		\begin{subfigure}{0.24\textwidth}
			\includegraphics[trim={0 5 43 35},clip,width=\linewidth]{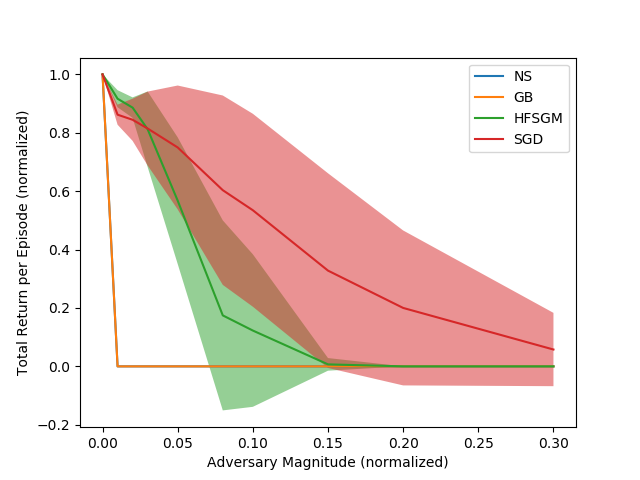} 
			\caption{DDQN Mountain Car}
			%		\label{fig:rew_cart}
		\end{subfigure} 
		\begin{subfigure}{0.24\textwidth}
			\includegraphics[trim={0 5 43 35},clip,width=\linewidth]{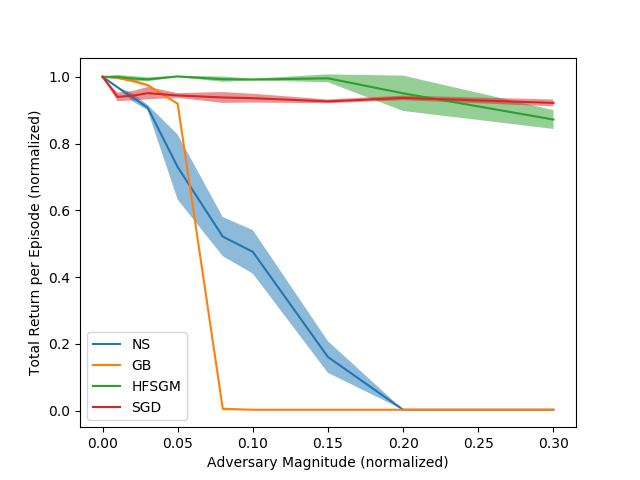}
			\caption{RBF Q mountain car}
			%		\label{fig:rew_mouncar}
		\end{subfigure}
	\end{center}	
	\caption{Comparison of different attacks on DDQN and RBF based Q learning. Subfigure (a) and (b) shows adversarial attack on cart-pole environment with DDQN. Subfigures (c) and (d) show adversarial attack on mountain car environment. It can be observed that Gradient Based (GB) attack performs better than Naive Sampling (NS) which in turn outperform Stochastic Gradient Descent (SGD) as well as HFSGM (\cite{huang2017adversarial}).  RBF Q learning is relatively more resilient to adversarial attack than DDQN.  
		% It is clear that the naive adversarial attack has significantly degraded the performance of DQN while there is no significant degradation in performance of RBF based Q learning
	}
	\label{fig:att}
	
\end{figure*}
\subsection{Adversarial Attack}\label{subsec:result_attack}
We show that the proposed attack(s) outperform attacks in \cite{huang2017adversarial} as shown in Fig. \ref{fig:att}. Here NS refers to naive sampling attack, GB refers to gradient based attack, HFSGM refers to the attack in \cite{huang2017adversarial} and SGD refers to the stochastic gradient descent attack. In Fig. \ref{fig:att}, the rewards as well as magnitude of adversarial attacks have been normalized to 1 where reward of 1 corresponds to the average reward received without adversarial attack and adversary magnitude of 1 corresponds to maximum possible value of the state. As we can observe from Fig. \ref{fig:att}, the adversarial attacks have degraded the performance of deep learning based algorithms. Furthermore, naive sampling based attacks and gradient based attacks perform better FSGM and SGD. Amongst naive sampling and gradient based attack, gradient based attack performs better. Another observation is the relative resilience of RBF based Q learning algorithm as compared to DDQN. This can be explained by the fact that Deep Networks might be providing ``jerky'' or piecewise linear function approximators as opposed to smoother interpolation of RBF. Thus, a small perturbation in state causes agent to take bad action. The piecewise linear function approximation of DNN can be attributed to the use of ``popular'' ReLu (or leaky ReLu) activation function as the output is ultimately a composition of piecewise linear functions.     
%\subsection{Architeture for Deep Deterministic Policy Gradient (DDPG)}
%For DDPG, the actor network consisted of 3 layers of 32 hidden units and the output was action. The activation function for all the hidden units were ReLu. The exploration was determined by Ornstein-Uhlenbeck process  with $\mu = 0$ and $\sigma= 0.3$ and $\theta = 0.5$. The critic layer consisted of the observation and action as input, 2 layers of 16 units with ReLu activation function were used as hidden layers. The adversarial noise that was induced here was uniform noise whose value of support was around 5\% the values taken by observations in the corresponding state.

%\subsection{Architecture for Radial Basis Function based Actor-Critic}
%For RBF based actor critic, the actor consisted of 64 nodes with Gaussian kernels. Again, the centres were scattered uniformly over the observation space. The critic network consisted of 400 nodes of Gaussian kernels with centres scattered uniformy over action space. The adversarial noise was identical to the one for DDPG.

\subsection{Robust Training}\label{subsec:robust_train_res}
\begin{figure}
	
	\begin{subfigure}{0.24\textwidth}
		\includegraphics[trim={70 0 50 25},clip, width=\linewidth]{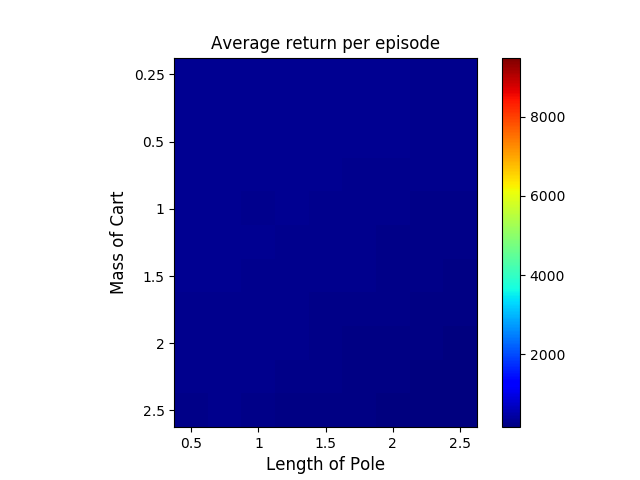} 
		\caption{DDQN Cartpole}
		\label{fig:ddqn_cart}
	\end{subfigure}
	\begin{subfigure}{0.24\textwidth}
		\includegraphics[trim={70 0 50 25},clip, width=\linewidth]{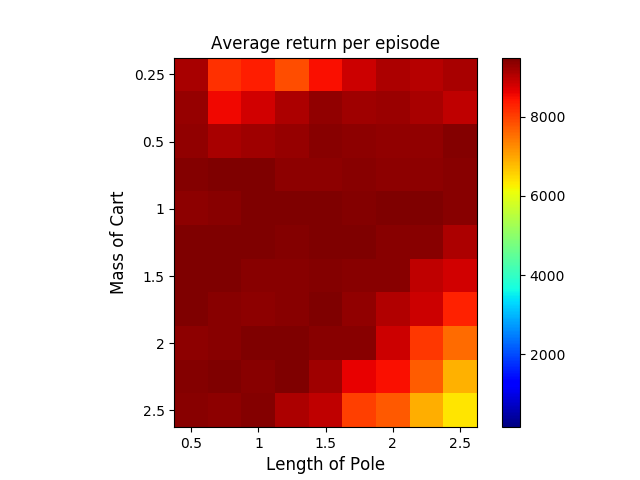}
		\caption{\centering{Robust DDQN Cartpole}}
		\label{fig:rob_ddqn_cart}
	\end{subfigure}
	%	\begin{subfigure}{0.23\textwidth}
	%		\includegraphics[trim={40 0 40 25},clip, width=\linewidth]{figures/cartpole_train_diff.png}
%	\caption{Subfigure (a) shows the average return per episode for cart-pole environment using DDQN algorithm across variation of mass of cart and length of pole. Subfigure(b) shows the same information for adversarially trained DDQN agent. We can observe humongous improvement over the return for agent across different parameters. ``Zoomed" colormap for DDQN cartpole comparison is provided in supplemental material.}
%	\label{fig:train_cart}
	%	\end{subfigure}
	\begin{subfigure}{0.24\textwidth}
		\includegraphics[trim={30 0 40 25},clip, width=\linewidth]{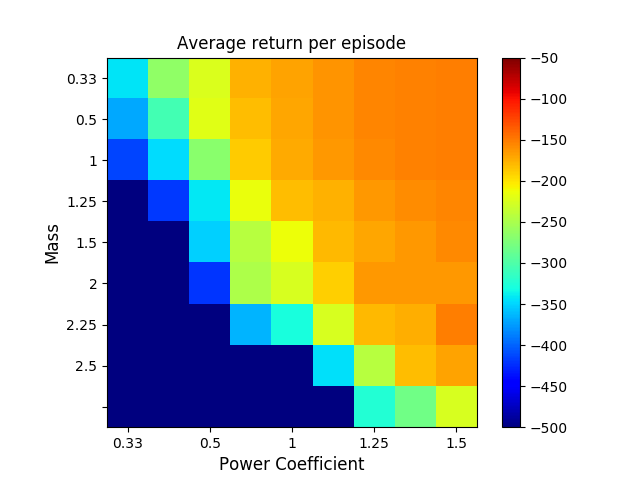}
		\caption{DDQN Mountain Car}
		\label{fig:ddqn_moun}
	\end{subfigure}
	\begin{subfigure}{0.24\textwidth}
		\includegraphics[trim={30 0 40 25},clip, width=\linewidth]{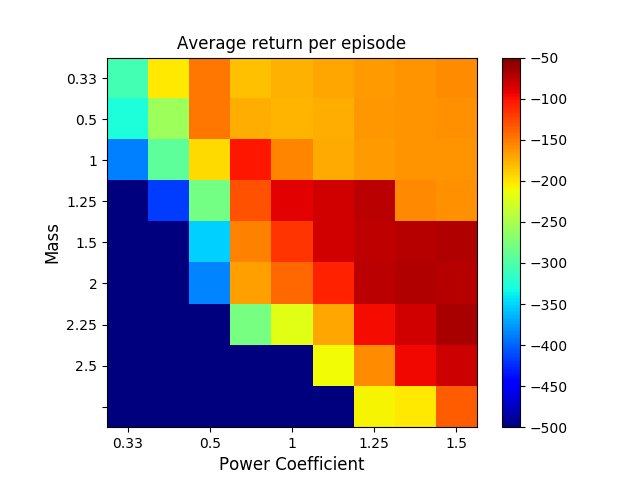}
		\caption{\centering{Robust DDQN MountainCar}}
		\label{fig:rob_ddqn_moun}
	\end{subfigure}
	%	\begin{subfigure}{00.23\textwidth}
	%		\includegraphics[trim={20 0 40 25},clip, width=\linewidth]{figures/mouncar_train_diff.png}
%	\caption{Subfigure(a) shows the return of DDQN agent for mountain car environment across variation of mass of car and the power coefficient of the car. Subfigure(b) shows the same information for robust DDQN. We can observe improvement mostly across the diagonal. The reason has been discussed in subsection \ref{subsec:robust_train_res}.}
%	\label{fig:train_mouncar}
	%	\end{subfigure}
	\begin{subfigure}{00.24\textwidth}
		\includegraphics[trim={20 10 10 10},clip, width=\linewidth]{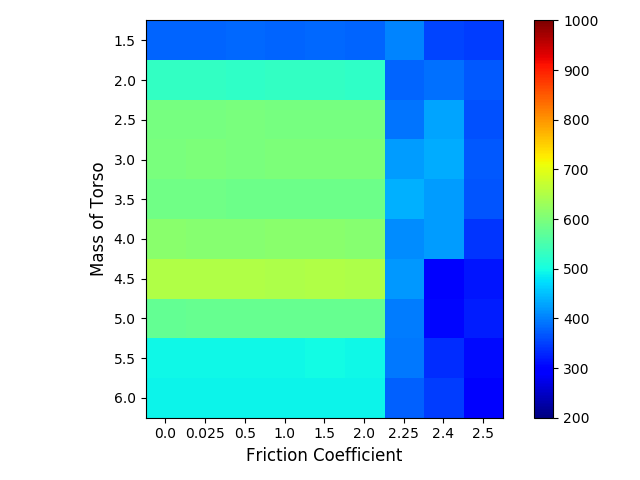}
		\caption{DDPG Hopper}
		\label{fig:ddpg_hop}
	\end{subfigure}	
	\begin{subfigure}{00.24\textwidth}
		\includegraphics[trim={20 10 10 10},clip, width=\linewidth]{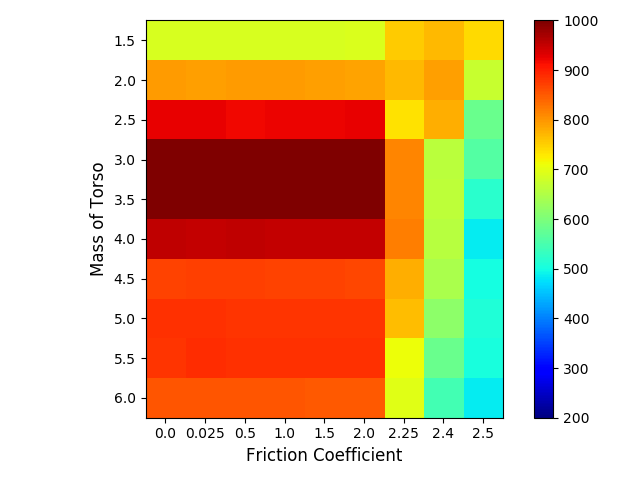}
		\caption{\centering{Robust DDPG Hopper}}
		\label{fig:rob_ddpg_hop}
	\end{subfigure}	
	%	\begin{subfigure}{00.23\textwidth}
	%	\includegraphics[trim={20 0 40 25},clip, width=\linewidth]{figures/hopper_diff.png}
%	\caption{Subfigure (a) shows the average return per episode for DDPG algorithm in hopper environment across variations of mass of torso of hopper and sliding friction coefficient with ground. Subfigure (b) shows the same information for robust DDPG. We can observe improvement across parameters because of robust training.   }
%	\label{fig:train_hopper}
	%	\end{subfigure}
	\begin{subfigure}{00.24\textwidth}
		\includegraphics[trim={10 10 10 35},clip, width=\linewidth]{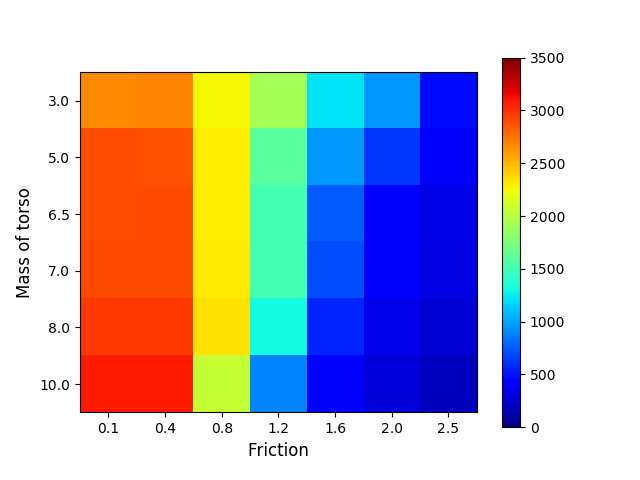}
		\caption{DDPG Half Cheetah}
		\label{fig:ddpg_che}
	\end{subfigure}
	\begin{subfigure}{00.24\textwidth}
		\includegraphics[trim={10 10 10 35},clip, width=\linewidth]{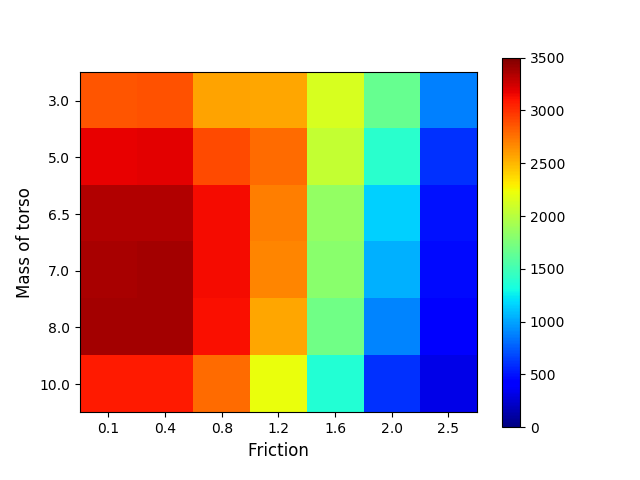}
		\caption{\centering{Robust DDPG Half Cheetah}}
		\label{fig:rob_ddpg_che}
	\end{subfigure}		
	%	\begin{subfigure}{00.23\textwidth}
	%	\includegraphics[trim={20 0 40 25},clip, width=\linewidth]{figures/che_diff.png}
	\caption{Subfigure (a) shows the average return per episode for DDQN algorithm in Cart-pole environment across variations of mass of cart and length of pole. Subfigure (b) shows the average return per episode of Robust DDQN algorithm. Similar results have been provided in Subfig. (c) and (d) for Mountain Car environment. We can observerve high return over wide range of parameters and over "default" parameter owing to robust training. Subfigures (e) and (f) show efficacy of robust training of DDPG algorithm. Subfigures (g) and (h) again demonstrate superior performance of Robust DDPG for Half-Cheetah environment.}
	\label{fig:robust_train}
\end{figure}

We present results that show significant improvement in robustness because of proposed adversarial training algorithm. For robust adversarial training, we first trained an agent with ``vanilla'' DRL (DDQN, DDPG) on ``default'' parameters. The trained agent is then made robust through adversarial training with gradient based adversarial attacks on same parameters. For evaluation, we tested this adversarially trained agent on a wide range of parameters and compared it to ``vanilla'' DRL. Figures \ref{fig:robust_train} shows improvement in performance owing to robust training. In Fig. \ref{fig:ddqn_cart}, \ref{fig:rob_ddqn_cart}, there is significant improvement in the overall reward for cartpole environment. The agent receives reward of 1 for each time step during which the cartpole is balanced. In  \ref{fig:ddqn_cart}, it seems that the agent receives uniform reward because of the scale used. A "zoomed" image of it is provided in Appendix. In mountain car environment (Fig. \ref{fig:ddqn_moun} and \ref{fig:rob_ddqn_moun}), the agent receives a negative reward of -1 for each timestep that it takes to reach the goal. The episode ends either when the car reaches goal or takes up 500 timesteps to reach the target. It is interesting to observe significant gains because of robust training around diagonal where the power coefficient is just enough to push the mountain car. In the lower triangular region, there is no improvement because the power coefficient is not enough for the mountain car and is underpowered. In the upper right region, it is overpowered. Hence, it reaches goal easily. For hopper environment (Fig. \ref{fig:ddpg_hop} and \ref{fig:rob_ddpg_hop}), higher return over DDPG can be attributed to adversarial training (robust DDPG). Similar results can also be observed for Half Cheetah environment (Fig. \ref{fig:ddpg_che} and \ref{fig:rob_ddpg_che}). These results show average return of 100 episodes for each set of parameter variation over 4 different seeds. The number of training steps is same for both vanilla algorithms and adversarially trained algorithms. Thus, the improvement is not because of more training.
\section{Conclusion}{\label{sec:conclusion}}
In this paper, we have proposed adversarial attack for reinforcement learning algorithms. We show that DRL can be fooled easily as compared to reinforcement learning algorithms based on Radial Basis Function (RBF) network. Interestingly, naive adversarial attacks on DRL can degrade it as opposed to robust policy learnt by RBF. We leveraged these attacks to train RL agent that led to robust performance across parameter variations for DDPG and DDQN. Future direction involves providing theoretical relationship between these attacks and robustness of the algorithms (to parameter variation).

%===============================================================================

%\bibliographystyle{aaai}\bibliography{daslab_pubs,daslab_all}  % .bib

%\bibliographystyle{aaai}\bibliography{daslab_pubs,daslab_all}  % .bib
\bibliographystyle{plainnat}
\bibliography{daslab_pubs,daslab_all}

\begin{thebibliography}{26}
\providecommand{\natexlab}[1]{#1}
\providecommand{\url}[1]{\texttt{#1}}
\expandafter\ifx\csname urlstyle\endcsname\relax
  \providecommand{\doi}[1]{doi: #1}\else
  \providecommand{\doi}{doi: \begingroup \urlstyle{rm}\Url}\fi

\bibitem[Brockman et~al.(2016)Brockman, Cheung, Pettersson, Schneider,
  Schulman, Tang, and Zaremba]{brockman2016openai}
Greg Brockman, Vicki Cheung, Ludwig Pettersson, Jonas Schneider, John Schulman,
  Jie Tang, and Wojciech Zaremba.
\newblock Openai gym.
\newblock \emph{arXiv preprint arXiv:1606.01540}, 2016.

\bibitem[Carlini and Wagner(2016)]{carlini2016towards}
Nicholas Carlini and David Wagner.
\newblock Towards evaluating the robustness of neural networks.
\newblock \emph{arXiv preprint arXiv:1608.04644}, 2016.

\bibitem[Duan et~al.(2016)Duan, Chen, Houthooft, Schulman, and
  Abbeel]{duan2016benchmarking}
Yan Duan, Xi~Chen, Rein Houthooft, John Schulman, and Pieter Abbeel.
\newblock Benchmarking deep reinforcement learning for continuous control.
\newblock In \emph{International Conference on Machine Learning}, pages
  1329--1338, 2016.

\bibitem[Garc{\'\i}a and Fern{\'a}ndez(2015)]{garcia2015comprehensive}
Javier Garc{\'\i}a and Fernando Fern{\'a}ndez.
\newblock A comprehensive survey on safe reinforcement learning.
\newblock \emph{Journal of Machine Learning Research}, 16:\penalty0 1437--1480,
  2015.

\bibitem[Geibel and Wysotzki(2005)]{Geibel:2005:RRL:1622519.1622522}
Peter Geibel and Fritz Wysotzki.
\newblock Risk-sensitive reinforcement learning applied to control under
  constraints.
\newblock \emph{J. Artif. Int. Res.}, 24\penalty0 (1):\penalty0 81--108, July
  2005.
\newblock ISSN 1076-9757.
\newblock URL \url{http://dl.acm.org/citation.cfm?id=1622519.1622522}.

\bibitem[Geramifard et~al.(2013)Geramifard, Walsh, Tellex, Chowdhary, Roy, and
  How]{Geramifard:2013:TLF:2688182.2688183}
Alborz Geramifard, Thomas~J. Walsh, Stefanie Tellex, Girish Chowdhary, Nicholas
  Roy, and Jonathan~P. How.
\newblock A tutorial on linear function approximators for dynamic programming
  and reinforcement learning.
\newblock \emph{Found. Trends Mach. Learn.}, 6\penalty0 (4):\penalty0 375--451,
  December 2013.
\newblock ISSN 1935-8237.
\newblock \doi{10.1561/2200000042}.
\newblock URL \url{http://dx.doi.org/10.1561/2200000042}.

\bibitem[Goodfellow et~al.(2014)Goodfellow, Shlens, and
  Szegedy]{goodfellow2014explaining}
Ian~J Goodfellow, Jonathon Shlens, and Christian Szegedy.
\newblock Explaining and harnessing adversarial examples.
\newblock \emph{arXiv preprint arXiv:1412.6572}, 2014.

\bibitem[Gosavi(2014)]{gosavi2014variance}
Abhijit Gosavi.
\newblock Variance-penalized markov decision processes: Dynamic programming and
  reinforcement learning techniques.
\newblock \emph{International Journal of General Systems}, 43\penalty0
  (6):\penalty0 649--669, 2014.

\bibitem[Huang et~al.(2017)Huang, Papernot, Goodfellow, Duan, and
  Abbeel]{huang2017adversarial}
Sandy Huang, Nicolas Papernot, Ian Goodfellow, Yan Duan, and Pieter Abbeel.
\newblock Adversarial attacks on neural network policies.
\newblock \emph{arXiv preprint arXiv:1702.02284}, 2017.

\bibitem[Kos and Song(2017)]{kos2017delving}
Jernej Kos and Dawn Song.
\newblock Delving into adversarial attacks on deep policies.
\newblock \emph{arXiv preprint arXiv:1705.06452}, 2017.

\bibitem[Levine et~al.(2016)Levine, Finn, Darrell, and Abbeel]{levine2016end}
Sergey Levine, Chelsea Finn, Trevor Darrell, and Pieter Abbeel.
\newblock End-to-end training of deep visuomotor policies.
\newblock \emph{Journal of Machine Learning Research}, 17\penalty0
  (39):\penalty0 1--40, 2016.

\bibitem[Lillicrap et~al.(2015)Lillicrap, Hunt, Pritzel, Heess, Erez, Tassa,
  Silver, and Wierstra]{lillicrap2015continuous}
Timothy~P Lillicrap, Jonathan~J Hunt, Alexander Pritzel, Nicolas Heess, Tom
  Erez, Yuval Tassa, David Silver, and Daan Wierstra.
\newblock Continuous control with deep reinforcement learning.
\newblock \emph{arXiv preprint arXiv:1509.02971}, 2015.

\bibitem[Lin et~al.(2017)Lin, Hong, Liao, Shih, Liu, and Sun]{lin2017tactics}
Yen-Chen Lin, Zhang-Wei Hong, Yuan-Hong Liao, Meng-Li Shih, Ming-Yu Liu, and
  Min Sun.
\newblock Tactics of adversarial attack on deep reinforcement learning agents.
\newblock \emph{arXiv preprint arXiv:1703.06748}, 2017.

\bibitem[Mandlekar et~al.(2017)Mandlekar, Zhu, Garg, Li, and
  Savarese]{mandlekar2017iros}
Ajay Mandlekar, Yuke Zhu, Animesh Garg, Fei-Fei Li, and Silvio Savarese.
\newblock Adversarially robust policy learning: Active construction of
  physically-plausible perturbations.
\newblock \emph{IEEE International Conference on Intelligent Robots and Systems
  (to appear)}, 2017.

\bibitem[Mnih et~al.(2015)Mnih, Kavukcuoglu, Silver, Rusu, Veness, Bellemare,
  Graves, Riedmiller, Fidjeland, Ostrovski, Petersen, Beattie, Sadik,
  Antonoglou, King, Kumaran, Wierstra, Legg, and Hassabis]{mnih2015human}
Volodymyr Mnih, Koray Kavukcuoglu, David Silver, Andrei~A. Rusu, Joel Veness,
  Marc~G. Bellemare, Alex Graves, Martin Riedmiller, Andreas~K. Fidjeland,
  Georg Ostrovski, Stig Petersen, Charles Beattie, Amir Sadik, Ioannis
  Antonoglou, Helen King, Dharshan Kumaran, Daan Wierstra, Shane Legg, and
  Demis Hassabis.
\newblock Human-level control through deep reinforcement learning.
\newblock \emph{Nature}, 518\penalty0 (7540):\penalty0 529--533, 2015.

\bibitem[Mnih et~al.(2016)Mnih, Badia, Mirza, Graves, Lillicrap, Harley,
  Silver, and Kavukcuoglu]{mnih2016asynchronous}
Volodymyr Mnih, Adria~Puigdomenech Badia, Mehdi Mirza, Alex Graves, Timothy~P
  Lillicrap, Tim Harley, David Silver, and Koray Kavukcuoglu.
\newblock Asynchronous methods for deep reinforcement learning.
\newblock In \emph{International Conference on Machine Learning}, 2016.

\bibitem[Morimoto and Doya(2005)]{morimoto2005robust}
Jun Morimoto and Kenji Doya.
\newblock Robust reinforcement learning.
\newblock \emph{Neural computation}, 17\penalty0 (2):\penalty0 335--359, 2005.

\bibitem[Papernot et~al.(2016)Papernot, McDaniel, Wu, Jha, and
  Swami]{papernot2016distillation}
Nicolas Papernot, Patrick McDaniel, Xi~Wu, Somesh Jha, and Ananthram Swami.
\newblock Distillation as a defense to adversarial perturbations against deep
  neural networks.
\newblock In \emph{Security and Privacy (SP), 2016 IEEE Symposium on}, pages
  582--597. IEEE, 2016.

\bibitem[Pinto et~al.(2017)Pinto, Davidson, Sukthankar, and
  Gupta]{pinto2017robust}
Lerrel Pinto, James Davidson, Rahul Sukthankar, and Abhinav Gupta.
\newblock Robust adversarial reinforcement learning.
\newblock \emph{arXiv preprint arXiv:1703.02702}, 2017.

\bibitem[Plappert(2016)]{plappert2016kerasrl}
Matthias Plappert.
\newblock keras-rl.
\newblock \url{https://github.com/matthiasplappert/keras-rl}, 2016.

\bibitem[Rajeswaran et~al.(2016)Rajeswaran, Ghotra, Ravindran, and
  Levine]{rajeswaran2016epopt}
Aravind Rajeswaran, Sarvjeet Ghotra, Balaraman Ravindran, and Sergey Levine.
\newblock Epopt: Learning robust neural network policies using model ensembles.
\newblock \emph{arXiv preprint arXiv:1610.01283}, 2016.

\bibitem[Schulman et~al.(2015)Schulman, Levine, Abbeel, Jordan, and
  Moritz]{schulman2015trust}
John Schulman, Sergey Levine, Pieter Abbeel, Michael~I Jordan, and Philipp
  Moritz.
\newblock Trust region policy optimization.
\newblock In \emph{ICML}, pages 1889--1897, 2015.

\bibitem[Silver et~al.(2016)Silver, Huang, Maddison, Guez, Sifre, Van
  Den~Driessche, Schrittwieser, Antonoglou, Panneershelvam, Lanctot, Dieleman,
  Grewe, Nham, Kalchbrenner, Sutskever, Lillicrap, Leach, Kavukcuoglu, Graepel,
  and Hassabis]{silver2016mastering}
David Silver, Aja Huang, Chris~J Maddison, Arthur Guez, Laurent Sifre, George
  Van Den~Driessche, Julian Schrittwieser, Ioannis Antonoglou, Veda
  Panneershelvam, Marc Lanctot, Sander Dieleman, Dominik Grewe, John Nham, Nal
  Kalchbrenner, Ilya Sutskever, Timothy Lillicrap, Madeleine Leach, Koray
  Kavukcuoglu, Thore Graepel, and Demis Hassabis.
\newblock Mastering the game of go with deep neural networks and tree search.
\newblock \emph{Nature}, 529\penalty0 (7587):\penalty0 484--489, 2016.

\bibitem[Tamar et~al.(2015)Tamar, Glassner, and Mannor]{tamar2015optimizing}
Aviv Tamar, Yonatan Glassner, and Shie Mannor.
\newblock Optimizing the cvar via sampling.
\newblock In \emph{AAAI}, pages 2993--2999, 2015.

\bibitem[Todorov et~al.(2012)Todorov, Erez, and Tassa]{todorov2012mujoco}
Emanuel Todorov, Tom Erez, and Yuval Tassa.
\newblock Mujoco: A physics engine for model-based control.
\newblock In \emph{IROS}, 2012.

\bibitem[Van~Hasselt et~al.(2016)Van~Hasselt, Guez, and Silver]{van2016deep}
Hado Van~Hasselt, Arthur Guez, and David Silver.
\newblock Deep reinforcement learning with double q-learning.
\newblock In \emph{AAAI}, pages 2094--2100, 2016.

\end{thebibliography}
\newpage
\appendix \begin{center} \huge{\textbf{APPENDIX}} \end{center}

\section{Experimental setup}
\subsection{DDQN}
The Deep Double Q learning for cart pole environment used 3 layers of 16 units each with Rectified Linear Unit (ReLu) activation function whereas the mountain car environment used 2 hidden layers of 100 units each of ReLu activation function. The discount factor for both of them was set at $0.99$and target network update rate were $10^{-2}$. The ``supervised learning" of networks was done with Adam optimization and learning rate of $10^{-3}$. The cartpole environment was trained for 50000 timesteps while Mountain Car was trained for 40000 time steps. The repository that we used was Keras-rl (\cite{plappert2016kerasrl}).
\subsection{RBF}
For Cart-pole, each dimension of state input was divided into 3 bins (b). The centroids were uniformly distributed along those bins. The variance of radial activation were $\frac{2}{b^2}$ . Discount factor of 0.99 was used. The learning rate was given by $0.001$. It was trained for 40000 time steps For Mountain car environment, the learning rate was $0.01$ and number of bins were 4. The discount factor was 0.99. Total number of timesteps were 60000
\subsection{DDPG}
For hopper and half cheetah environment, there were 2 hidden layers of 400 and 300 ReLu units for both actor and critic networks. The number of time steps it was trained were 1 million. Discount factor was 0.99. The learning rate of critic network was $10^{-3}$ while the learning rate of actor was $10^{-4}$. Half cheetah also used the same network as hopper. It also had the same learning rate and discount factor. It was trained for 2 million time steps. The repository that we used was rllab (\cite{duan2016benchmarking})
\subsection{Adversarial Training}
For adversarial training, the number of times sampling was done was 200 and the vanilla trained network was re-trained adversarially for same amount of time steps. The adversarial magnitude used was 0.05 for half cheetah and 0.03 for hopper. The sampling frequency 100. We must point out that for the results shown in paper, comparison has been shown between both vanilla and adversarially trained networks that have been trained for exactly same number of timesteps. 

\section{Robust Training Colormap for Cartpole}\label{subsec:robust_train_res}
\begin{figure}[!h]	
	\begin{subfigure}{0.35\textwidth}
		\includegraphics[trim={40 0 40 25},clip, width=\linewidth]{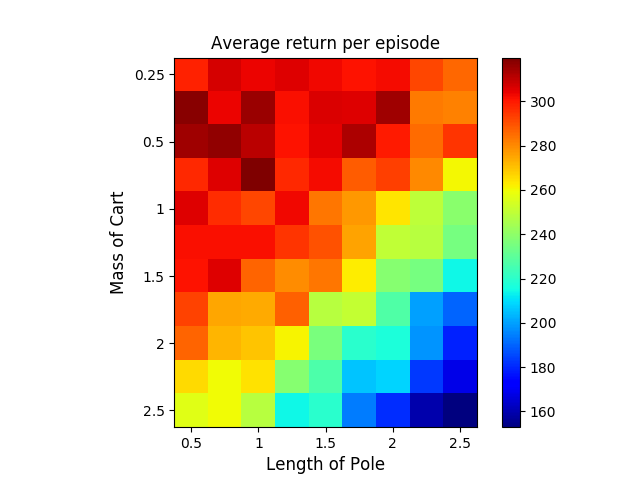} 
		\caption{DDQN Cartpole}
		%		\label{fig:ddpg}
	\end{subfigure}
	\begin{subfigure}{0.35\textwidth}
		\includegraphics[trim={40 0 40 25},clip, width=\linewidth]{figures/cartpole_train_adv.png}
		\caption{Robust DDQN Cartpole}
		%		\label{fig:rbfActorCritic}
	\end{subfigure}
	%	\begin{subfigure}{0.23\textwidth}
	%		\includegraphics[trim={40 0 40 25},clip, width=\linewidth]{figures/cartpole_train_diff.png}
	\caption{Subfigure (a) shows the average return per episode for cart-pole environment using DDQN algorithm across variation of mass of cart and length of pole. Subfigure(b) shows the same information for adversarially trained DDQN agent. We can observe significant improvement over the return for agent across different parameters. ``Zoomed" colormap for DDQN cartpole comparison has been presented.}
	\label{fig:train_cart}
	%	\end{subfigure}
\end{figure}
\end{document}